\documentclass{article}
\usepackage{authblk}
\usepackage{fullpage}
\usepackage{makecell}
\usepackage{hyperref}
\usepackage{graphicx}
\usepackage{xspace}
\usepackage{array}
\usepackage{multirow}
\usepackage{makecell}
\usepackage{xr}
\usepackage{subcaption}
\usepackage{titling}
\usepackage[normalem]{ulem}

\begin{document}
\title{Dynamic Pooling Improves Nanopore Base Calling Accuracy}

\author{Vladim\'ir Bo\v{z}a, Peter Pere\v{s}\'ini, Bro\v{n}a Brejov\'a, Tom\'a\v{s} Vina\v{r}}
\affil{Faculty of Mathematics, Physics and Informatics, Comenius University in Bratislava,\\ Mlynsk\'a dolina, 842 48 Bratislava, Slovakia}
\date{}
\maketitle

\begin{abstract}
In nanopore sequencing, electrical signal is measured as DNA molecules pass through the sequencing pores.
  Translating these signals into DNA bases (base calling) is a highly non-trivial task, and its quality has a large impact on the sequencing accuracy.
  The most successful nanopore base callers to date use convolutional neural networks (CNN) to accomplish the task.

Convolutional layers in CNNs are typically composed of filters with constant window size, performing best in analysis of signals with uniform speed.
  However, the speed of nanopore sequencing varies greatly both within reads and between sequencing runs.
  Here, we present dynamic pooling, a novel neural network component, which addresses this problem by adaptively adjusting the pooling ratio.
  To demonstrate the usefulness of dynamic pooling, we developed two base callers: Heron and Osprey.
  Heron improves the accuracy beyond the experimental high-accuracy base caller Bonito developed by Oxford Nanopore.
  Osprey is a fast base caller that can compete in accuracy with Guppy high-accuracy mode, but does not require GPU acceleration and achieves a near real-time speed on common desktop CPUs.

Availability: \href{https://github.com/fmfi-compbio/osprey}{\detokenize{https://github.com/fmfi-compbio/osprey}}, \href{https://github.com/fmfi-compbio/heron}{\detokenize{https://github.com/fmfi-compbio/heron}}


\end{abstract}

\section{Introduction}

The MinION by Oxford Nanopore Technologies (ONT) is a portable DNA sequencer, which is capable of producing very long reads \cite{leggett2017world, bayley2015nanopore}.
  The main drawback of this sequencing technology is its high error rate - the median read accuracy barely achieves 96\% (Table \ref{tab:acc}, see also \cite{silvestre2020pair}).
  The source of sequencing errors is the fact that the MinION measures tiny electrical current influenced by DNA passing through the pore.
  This noisy signal is then translated into DNA bases via base caller software. 

\begin{figure}
    \centering
  
    \includegraphics[width=0.8\textwidth]{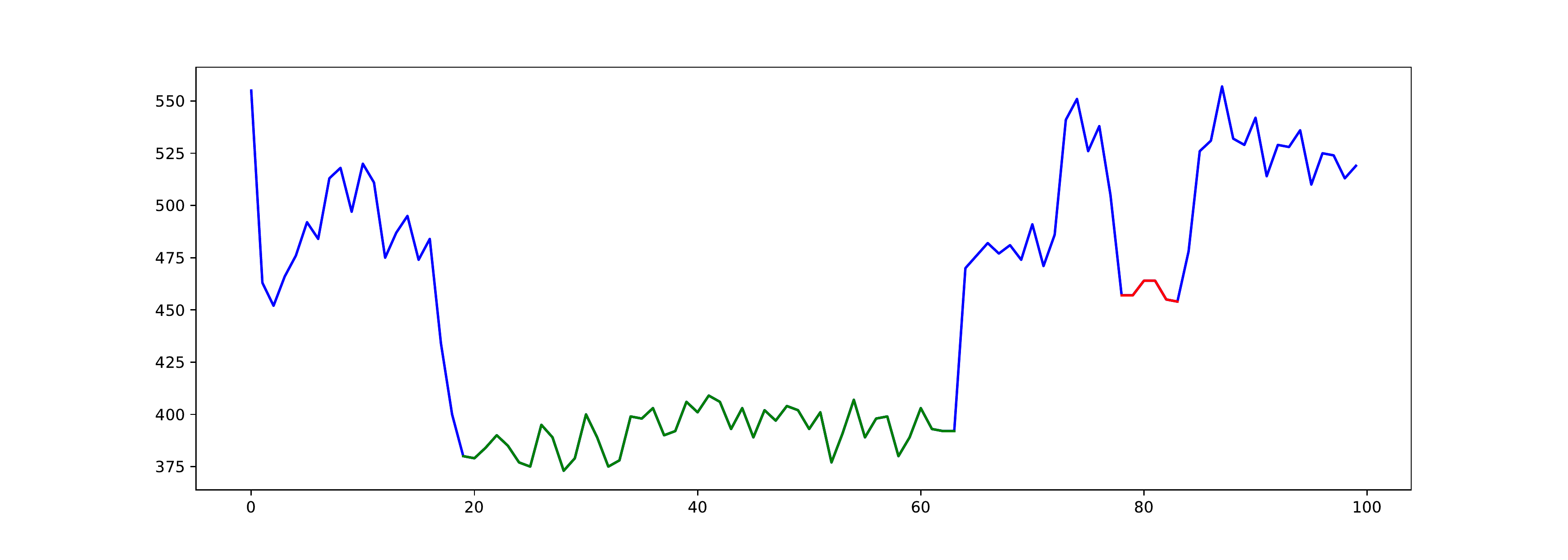}
    
    \caption{
      \emph{ \label{fig:signal} An example of nanopore signal, with two events highlighted in red and green, and the remaining signal shown in blue.
          This illustrates that the length of events can vary substantially.}
    }
  \end{figure}

The sequence of signal readouts can be split into \emph{events}, each event corresponding to a passage of a single DNA base through the pore. A single event consists of roughly 8-10 signal readouts on average.
  However, local variations in the speed of DNA passing through the pore cause the event length to vary widely, some events consisting of a single readout, while others spanning tens of readouts (see Figure \ref{fig:signal}).
  Moreover, the speed may also vary globally between the runs, making some runs on average slower than others.
  Segmentation of the signal into events is a difficult problem, and most base callers avoid performing explicit event segmentation.

Currently, the most successful base callers are based on neural networks, using either recurrent neural network architecture (Guppy, Bonito v0.3 \cite{bonito2020}, Chiron \cite{teng2018chiron}, Deepnano-blitz \cite{boza2020deepnano}) or convolutional neural networks (Bonito v0.2, WaveNano \cite{wang2018wavenano}, CausalCall \cite{zeng2020causalcall}, Deepnano-coral \cite{peresini2020nanopore}). There are also base callers based on self-attention (transformer) architecture \cite{huang2020sacall,lv2020end}.

Recurrent neural networks (RNNs) are designed specifically for processing sequence data, and thus may seem a natural choice for nanopore base calling.
  However, convolutional neural networks (CNNs) offer several benefits.
  First, each CNN layer can process the whole sequence in parallel compared to sequential processing in RNNs.
  Thus, CNNs can achieve high throughput.
  Second, CNNs offer many possibilities for architecture optimization.
  Typical RNNs can be tuned mainly by changing the layer type, and the number and sizes of layers, and there were very few successful improvements documented in literature.
  Compared to that, CNNs offer a much larger variety of choices, including skip connections \cite{he2016deep, huang2017densely}, normalization \cite{ioffe2015batch}, activation functions \cite{ramachandran2017searching}, convolution factorization \cite{szegedy2016rethinking, howard2017mobilenets}, etc. A similar shift from RNNs to CNNs can be observed in natural language processing and speech recognition, both of which were first dominated by RNNs \cite{conneau2016very,dauphin2017language,pratap2018wav2letter}.

A basic building block of CNNs is a convolutional layer, which in the one-dimensional case takes as an input a tensor $X$of dimensions $(T,\ C_{in})$ representing a data stream of length $T$, each data point containing $C_{in}$ values called \emph{channels} and outputs tensor $Y$ with dimensions $(T,\ C_{out})$.
  The convolution with kernel size $D$is described by \emph{kernel weights$W$}of dimensions $(C_{out},\ D,\ C_{in})$and \emph{bias} vector $B$of length $C_{out}$, both of which are trained.
  The convolution is computed as follows: $Y_{t,\ j}=\sum\limits_{0\leq d<D,\ {0\leq i}_{\ }<C_{in}}^{\ }X_{t+d,\ i_{\ }}W_{j,d,i}+B_{j_{\ }}$(to handle boundaries, the input tensor $X$ is first padded with $\lfloor\frac{D}{2}\rfloor$zeros at both ends).
  The length of the output stream can be reduced by applying \emph{stride $s$}, where only every $s$-th element of the output is computed.
  A simpler version of striding is pooling, where a fixed operation, such as maximum or mean, is applied instead of learned kernel weights on windows of $D$points with stride $s$\emph{.}

To make base calling faster and reduce signal redundancy, CNN base callers often subsample the input signal.
  For example, the first layer of the Bonito network uses stride of size 3 in version 0.2 and size 5 in version 0.3.
  Besides increasing the time and memory efficiency, this step allows the neural network layers to increase their effective receptive field (i.e., the part of the signal they see) without increasing kernel sizes of internal components.
  However, subsampling with a fixed-sized striding step may completely discard very short events, while longer events are still overrepresented.
  In addition, the widely variable event lengths make base calling hard for CNNs because each convolutional layer considers fixed-size windows.
  Under these conditions it may be difficult for the network to capture dependencies between signal readouts at specific distances in the DNA, as these have variable distance in the actual signal processed at individual layers of the network.
   In particular, long events decrease the effective receptive field in terms of the actual DNA context observed, while very short events may suffer from reduced representation after downsampling.

To address these problems and provide convolutional layers with a more stable context width, we introduce a novel component, which we call \emph{dynamic pooling}.
  This component can adaptively subsample the signal by variable rate depending on the input.
  Intuitively, we want the neural network to  predict the current speed of the DNA at each position and then downsample signal readouts based on this factor.
  Such adaptability can enable the network to keep short events, and downsample long events.

Slightly similar approaches are Deformable convolution \cite{dai2017deformable} and Dynamic convolutions \cite{wu2019pay}.
  Both of these approaches do not change the number of timesteps in the sequence, but instead of applying the same convolution for each timestep, they use a different convolution in every timestep.
  Deformable convolutions use the same convolution weights but alter input positions.
  Dynamic convolutions compute different convolution weights for every timestep.
  
Thus these approaches would keep long events unnecessarily long, whereas our approach shortens long events to a reasonable size.
  Also these approaches need custom CUDA kernels to be efficient on current hardware, whereas our approach relies only on simple primitives.

A very straightforward approach to adaptive subsampling would be to predict for each input data point, whether the network should use it or ignore it.
  This is similar to conditional computation \cite{bengio2015conditional} and leads to a non-continuous loss function.
  Such an approach therefore cannot be trained by standard gradient descent algorithms. In contrast, our dynamic pooling approach is continuous and differentiable almost everywhere.

To demonstrate benefits of dynamic pooling, we apply it to two strong baseline base caller networks.
  The first (named Heron) is an improved version of the high-accuracy Bonito v0.2 architecture, an experimental CNN base caller from ONT. Our improvements, including dynamic pooling, boost the accuracy by almost 2 percentage points on a rice dataset, decreasing the error rate by approximately 33\%.

Secondly, we develop a fast  base caller, Osprey, which on a common desktop CPU can process 1.7 million signals per second, nearly matching the sequencing speed of MinION.
   Osprey has a better accuracy than Guppy 3.4 in high accuracy mode, but unlike Guppy, Osprey can be run efficiently on systems lacking high-performance GPUs.

\section{Methods}

In this section, we first describe dynamic pooling, our novel method for subsampling the signal to adapt to its varying speed and to provide subsequent layers with a more stable context within their receptive field.
  In the second part, we summarize technical details of application of this approach to the base calling problem.

\subsection{Dynamic Pooling}

Dynamic pooling, our approach to adaptive subsampling, works as follows: Given the input sequence $x_1,x_2,...,x_n$, we first calculate three new sequences:
\begin{itemize}

  \item feature vector: $f_i=f(x_i)\in (0,1{)}^C$ ($C$ is the number of output channels)
  
  \item point importance: $w_i=w(x_i),\ w_i\in (0,1)$
  
  \item length factor: $m_i=m(x_i),\ m_i\in (0,1)$
  
\end{itemize}

Each of the functions $f,w,m$ can be any differentiable function, typically a single convolution or a network with several convolutional layers.

\begin{figure}
    \centering
  
    \includegraphics[width=0.8\textwidth]{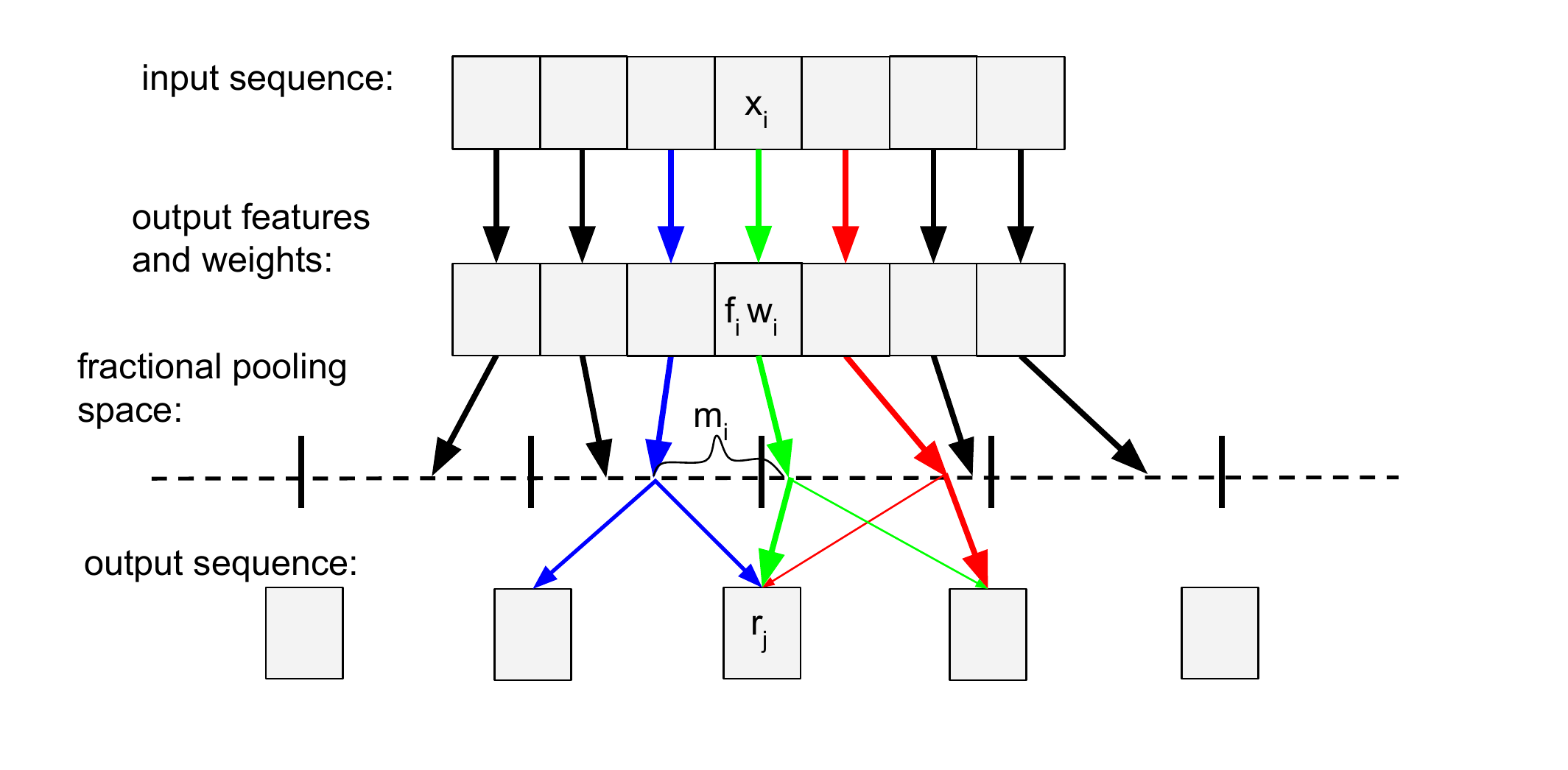}
    
    \caption{
      \emph{ \label{fig:dynpool} Illustration of the dynamic pooling scheme.
          Arrows show transformation of input features into weighted output features, their mapping into pooling space and summation into the final result.
          The arrow thickness in the last transformation represents the weight depending on the distance of a fractional point to the output point.}
    }
  \end{figure}

The main idea is that we time-warp the signal using length factors $m_i$ representing (fractional) distances of features in the pooling space (Figure \ref{fig:dynpool}).
  Each input point $x_i$ at position $i$ is thus mapped to a new position $p_i$ in the pooling space which is computed as the cumulative sum of the length factors up to index $i$:
$p_i=\sum\limits_{j\leq i}^{\ }m_j$

To produce the final output sequence of length $\lceil\ p_n\ \rceil$, we discretize the fractional signal positions and combine feature vectors that map to nearby positions by weighted nearest-neighbour resampling.
  That is, to compute output $r_j$ at position $j$, we compute a weighted sum of feature vectors which are mapped to pooling space between positions $j-1$ and $j+1$.
  The weights are given by the distance from $j$ and multiplied by the importance $w_i$.
  The resulting output vector $r_j$is thus computed as follows:
$r_j=\sum\limits_{i,\ {|p}_i-j|\leq 1}^{\ }f_i{w}_i(1\ -\ |p_i-j|)$

Note that our dynamic pooling layer is continuous and almost everywhere differentiable through $m$, $w$ and $f$.
  Also our formulation is a generalization of both mean pooling and striding.
  To obtain mean pooling with size $s$, we set $w_i=1/s$and $m_i=1$if $i=ks$ and $m_i=0$ otherwise.
  To obtain striding with step $s$,  we set $m_i=w_i=1$if $i=ks$ and $m_i=w_i=0$ otherwise.

Our experience indicates that several practical concerns need to be addressed in application of dynamic pooling.
  First, the average pooling factor can easily diverge during training if no bound on its size is set.
  Even though we saw correlation between the output size after pooling and the actual number of sequenced bases at some point of the training, the training often degenerated into average pooling 1 (i.e.
  no pooling at all) or  pooling so high that the output size was smaller than the number of sequenced bases.
  
To address this issue, as well as to roughly fix the target pooling factor (and thus computational budget), we borrow an idea from batch normalization \cite{ioffe2015batch}.
  Ideally, we would like the average pooling factor over the whole dataset to be equal to some fixed constant $S$.
  Since this is hard to achieve, we approximate this by keeping the average pooling factor fixed for each batch during training.
  We first calculate length factors ${m^{(s)}}_i$ for each sample $s$ and sequence point $i$, then we calculate the average length factor for the batch $M=\frac{1}{bn}\sum\limits_{s,i}^{\ }{m^{(s)}}_i$, and finally, we renormalize each length factor as $m{'^{(s)}}_i={m^{(s)}}_i\frac{S}{M}$.
  It is possible for the renormalized length factor to be greater than one, but we have not found any issues with that.
  Note that the pooling factor in each sequence may vary, but the average pooling factor per batch is fixed.
  We also keep the exponential moving average of normalization factors $\frac{S}{M}$, and use it during inference similarly to batch normalization.

We also saw issues with divergence during training.
  This was due to very high gradients in length factors.
  Notice that in the prefix sum, the gradient at point $j$ is the sum of gradients for positions between $j$ and $n$: $\frac{\partial L}{\partial m_j}=\sum\limits_{k=j}^n\frac{\partial L}{{\partial p}_k}$.
  This means that if there is any noise in gradients, all of it will be concentrated in $\frac{\partial L}{\partial m_1}$, which will have very huge variance.
  Also note that variance of gradient is different for every $\frac{\partial L}{\partial m_j}$.
  To fix these issues, we limit the gradient to pass only through 20 elements, and thus we approximate $\frac{\partial L}{\partial m_j}\approx \sum\limits_{k=j}^{j+20}\frac{\partial L}{{\partial p}_k}$

\subsection{Application to Base Calling}

\begin{figure}
    \centering
  
    \includegraphics[width=1.0\textwidth]{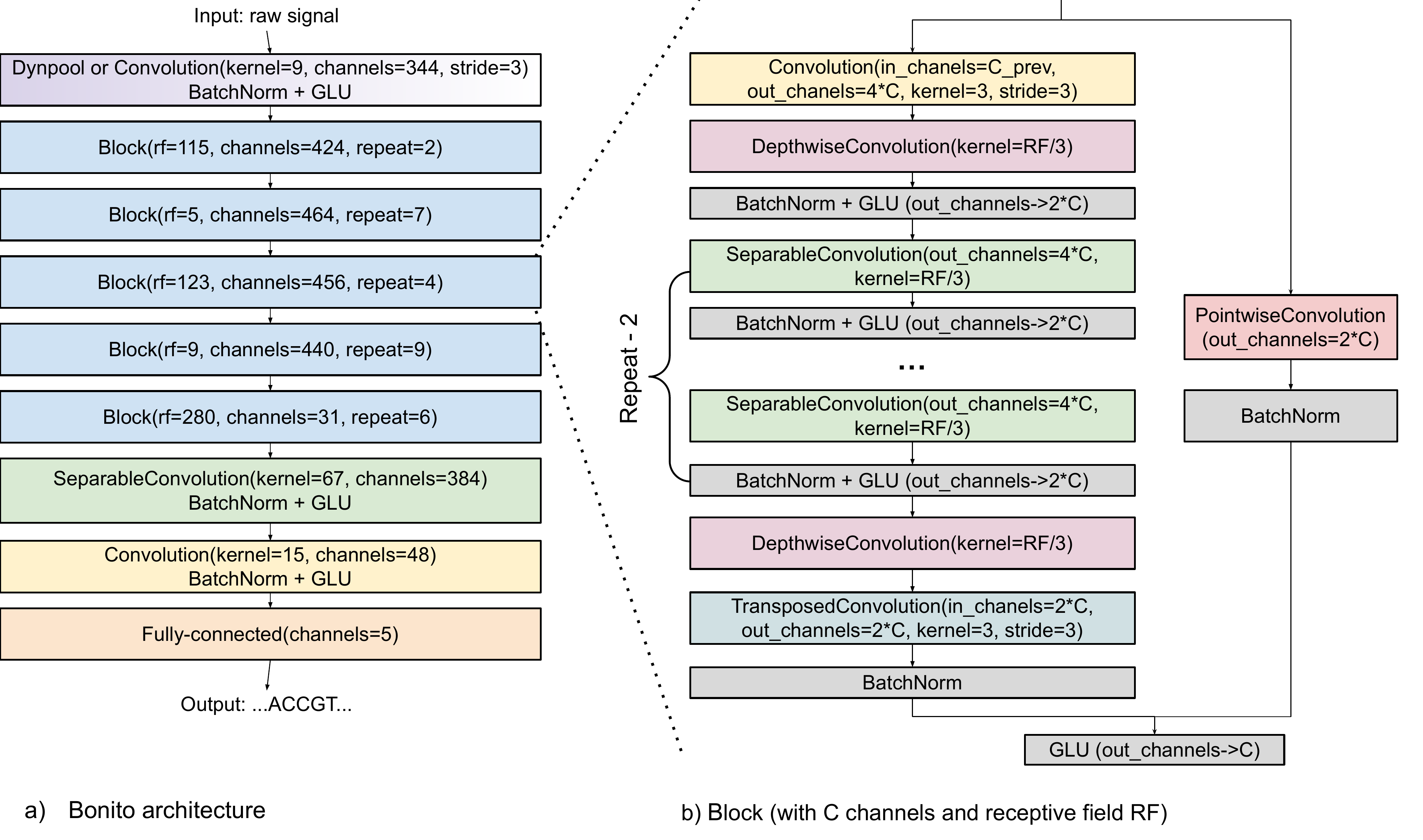}
    
    \caption{
      \emph{ \label{fig:hac_arch} High accuracy architecture.
          Block with gradient background represents block replaced by dynamic pooling.}
    }
  \end{figure}

\begin{figure}
    \centering
  
    \includegraphics[width=1.0\textwidth]{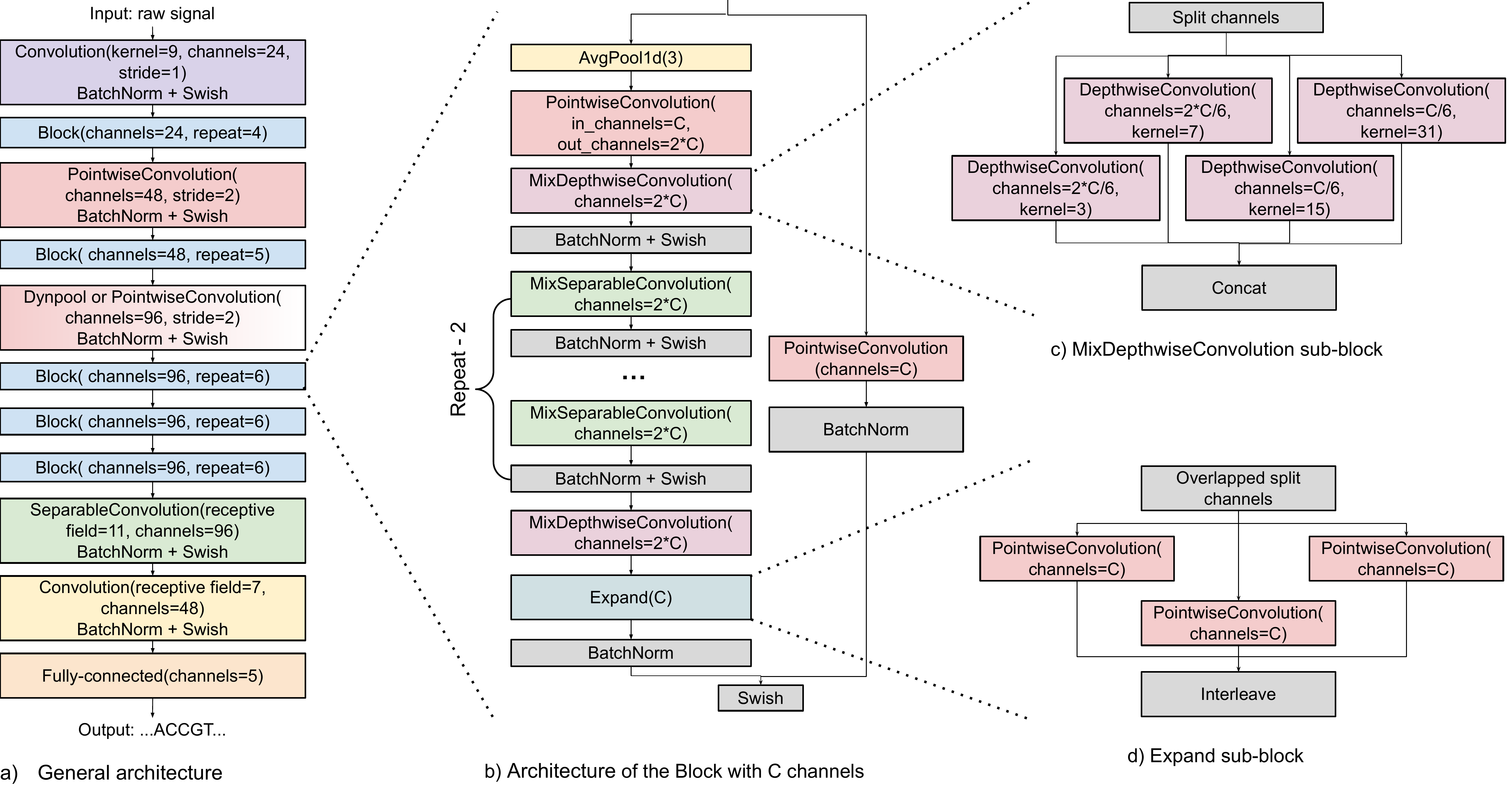}
    
    \caption{
      \emph{ \label{fig:fast_arch} Fast basecaller architecture.
          Block with gradient background represents block replaced by dynamic pooling}
    }
  \end{figure}

We developed two base callers, which use dynamic pooling to improve the speed vs.
  accuracy tradeoff.
  Heron is a high accuracy base caller, while Osprey is a base caller that runs in real time.
  Both base callers use deep convolutional neural networks inspired by Bonito v0.2, which is in turn based on the Quartznet speech recognition architecture \cite{kriman2020quartznet}.
  The architecture overview is shown in Figures \ref{fig:hac_arch} (Heron) and \ref{fig:fast_arch} (Osprey).

The basic building unit of these networks is the convolutional layer, described in the introduction.
  The networks use several special cases of convolutions.
  A pointwise convolution is a convolution with kernel size one (there is no interaction between neighbouring time points).
  A depthwise convolution is a convolution with no interaction between channels (the relevant parts of the weight tensor are diagonal matrices).
   Thus a depthwise convolution needs fewer parameters and fewer multiplications to compute.
  A depthwise convolution followed by a pointwise convolution is called a separable convolution.

Convolutional layers are organized into blocks.
  One block consists of a repeated application of a convolutional layer, batch normalization (\cite{ioffe2015batch}), and non-linearity.
  To allow smooth flow of gradient, blocks also employ skip connections, where the input of the block is processed by a single pointwise convolution followed by batch normalization and it is then added to the output.

In the rest of this section, we describe several improvements added to our networks;
    further details and ablation results can be found in the Supplementary material.

\textbf{Space-to-depth compression in blocks (both Heron and Osprey).
    }Peresini et al.
   \cite{peresini2020nanopore} have recently introduced this technique, where in each block, we compress the number of sequence points by a factor three and enlarge the number of channels by a factor two.
  This increases the number of trainable parameters, while only modestly increasing the total number of operations and running time.
  In Osprey, we limit the number of operations during compression and decompression by replacing the original convolution/transpose convolution with a mean pooling during compression and three smaller convolutions during decompression.

\textbf{Gated linear units (GLU) (Heron).
    }We change the original Swish nonlinearity \cite{ramachandran2017searching} to GLU \cite{dauphin2017language}.
  GLU first splits channels into two halves ($x_1,x_2$)  and then computes $g(x_1,x_2)=x_1\sigma (x_2)$.
  Note that Swish is a special case of GLU non-linearity with $x_1=x_2$.
  Even though the GLU comes with the cost of double the computation budget, in our context it works much better than increasing the number of channels in the network.

\textbf{Variable receptive fields in depthwise convolutions (Osprey).
    }Increasing the receptive field of convolution is often beneficial, but it comes at an increased computational cost.
  As a tradeoff, we split the channels into four groups in ratio 2:2:1:1 and apply a different receptive field in each group (3, 7, 15, and 31, respectively).

\textbf{Recurrent neural aligner as the output layer (Heron and Osprey).
    }Existing base callers (like Chiron, Deepnano-blitz, Bonito 0.2) produce an intermediate output with a simple distribution over five possibilities (A, C, G, T, or blank) for each sequence point.
  The subsequent connectionist temporal classification layer \cite{graves2006connectionist} produces the final sequence without blanks, selecting the one with the highest likelihood in the intermediate probability distribution.
  However, the intermediate distribution assumes that point predictions are  independent from each other.
  The recurrent neural aligner \cite{sak2017recurrent} introduces dependency between predictions by propagating state information.
  We describe this technique and its application to base calling in more detail in the Supplementary material, as we believe it may be useful also in other bioinformatics contexts.

\textbf{Details of dynamic pooling (Heron and Osprey).
    }We replace one of the strided convolutions with a dynamic pooling module (the exact application differs in Heron and Osprey, see Figures \ref{fig:fast_arch} and \ref{fig:hac_arch}).
   We calculate feature vector $f_i\ $using a one-layer convolution with the same number of channels and kernel as in the layer replaced by dynamic pooling (we set striding to one).
  For calculating moves and weights, we use a three-layer convolutional network in Heron and a simple one-layer pointwise convolution in Osprey.
  We use a deeper network in Heron, since we are operating on the raw signal.

\subsection{Datasets}

For training and testing of our base callers, we use a mix of samples from various datasets listed in Tables \ref{tab:train_datasets} and \ref{tab:testing_datasets}.

\begin{table*}
  \centering

  \begin{tabular}{|c|c|c|}
    \hline  Organism		& Total number of reads		& Total number of signals \\
    \hline  \parbox{8cm}{\centering
        Taiyaki train dataset \\(consists of human, \emph{E.
          coli} and \emph{S.
          cerevisiae}) \cite{taiyaki2020}}		& 48762		& 2979M \\
    \hline  \emph{E.
          coli} \cite{ecoli2020}		& 2804		& 797M \\
    \hline  Human \cite{jain2018nanopore}		& 9385		& 1106M \\
    \hline  Tomato \cite{schmidt2017novo}		& 6799		& 601M \\
    \hline
  \end{tabular}
  
  \caption{
    \emph{ \label{tab:train_datasets} Training datasets.
        }
  }
\end{table*}

\begin{table*}
  \centering

  \begin{tabular}{|c|c|c|}
    \hline  Organism		& Total number of reads		& Total number of signals \\
    \hline  \emph{Klebsiella  pneumoniae} \cite{wick2019performance}		& 1788		& 356M \\
    \hline  Mouse \cite{gigante2019using}		& 2089		& 153M \\
    \hline  Rice \cite{choi2020nanopore}		& 1949		& 473M \\
    \hline  Zymoset \cite{zymo2020}		& 4000		& 993M \\
    \hline
  \end{tabular}
  
  \caption{
    \emph{ \label{tab:testing_datasets} Testing datasets.
        }
  }
\end{table*}

\section{Experiments}

We evaluate our base callers on several testing data sets.
  Each base called read is aligned to the corresponding reference by minimap2 \cite{li2016minimap}, and the accuracy is computed as the fraction of correctly called bases in the aligned portion of the read. The median accuracy over the whole testing set is reported in a corresponding table.
  In case of our base callers, the testing data sets did not overlap with the training data sets;
   this cannot be guaranteed for other base callers.

\subsection{High Accuracy Results}

\begin{figure}
    \centering
  
    \includegraphics[width=0.9\textwidth]{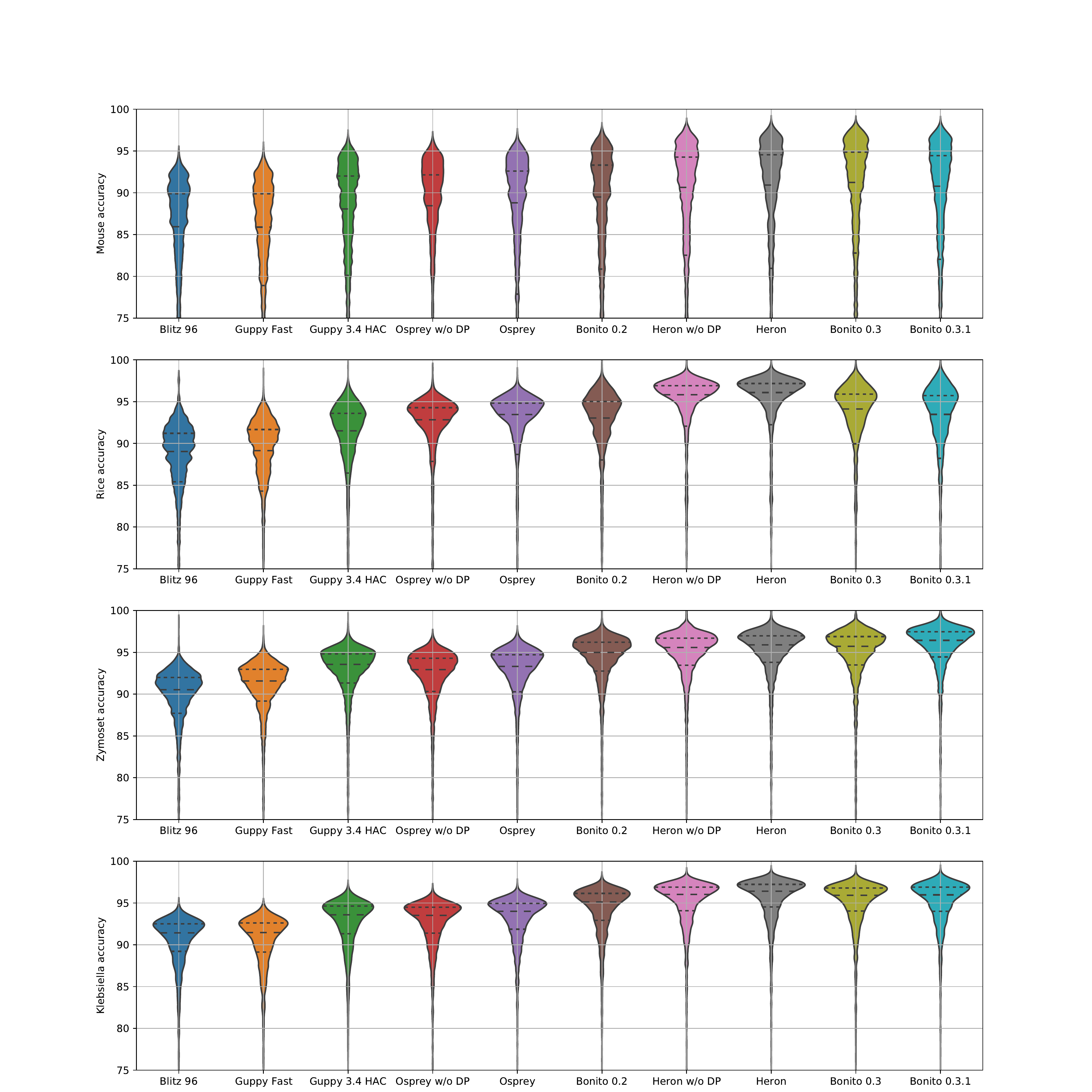}
    
    \caption{
      \emph{Figure: \label{fig:dist} Read accuracy distribution for individual testing sets (from top to bottom mouse, rice, zymoset, Klebsiella).}
    }
  \end{figure}

Heron is designed as a high accuracy base caller.
  As such, we compare its results to the series of experimental high accuracy base callers by Oxford Nanopore Technologies (Bonito 0.2, Bonito 0.3, and Bonito 0.3.1).
  Note that Bonito typically has a higher accuracy than the standard production base caller, Guppy in the high accuracy mode.
  Table \ref{tab:acc} and Figure \ref{fig:dist} show that Heron is faster than Bonito and achieves comparable accuracy, sometimes outperforming Bonito significantly.
  On the rice dataset, the error rate is reduced  by 40\% compared to Bonito 0.3.1.

\begin{table*}
  \centering

  \begin{tabular}{|c|c|c|c|c|c|}
    \hline  Base caller		& Mouse		& Rice 		& Zymoset		& \emph{Klebsiella}		& Speed [signals/s] \\
    \hline  Bonito 0.3.1		& 90.770		& 93.477		& \textbf{96.435}		& 95.981		& 0.37M \\
    \hline  Bonito 0.3		& \textbf{91.233}		&         94.129		& 95.733		& 95.914		& 0.37M \\
    \hline  Bonito 0.2		& 89.489		& 93.040		& 94.999		& 95.136		& 0.44M \\
    \hline  Heron w/o dynpooling		& 90.633		& 95.822		& 95.596		& 96.037		& 0.45M \\
    \hline  Heron with dynpooling		& 90.911		& \textbf{96.092}		& 95.907		& \textbf{96.407}		& 0.41M \\
    \hline
  \end{tabular}
  
  \caption{
    \emph{ \label{tab:acc} Median read accuracy of high accuracy base callers on several datasets.
        The speed is measured on T4 GPU.}
  }
\end{table*}

\subsection{Fast Base Calling Results}

Fast base callers are important to achieve live base calling necessary for applications such as run monitoring, selective sequencing etc.
  To achieve live base calling for MinION, one has to process approximately 1.5 million signals per second.
  Osprey has been designed to achieve this goal on a common four-core desktop CPU.
  Note that the only other base caller that can achieve the same goal on a CPU is Deepnano-blitz \cite{boza2020deepnano}.

Table \ref{tab:accfast} shows that Osprey is both more accurate and faster than the fast version of Guppy and can even compare in accuracy to a one year old high accuracy version of Guppy.

We have also investigated the high performance of Heron and Osprey on the rice dataset.
  We have noticed that in this dataset, the speed of DNA passing through the pore is slightly slower (Figure \ref{fig:dynpoolres}).
  We speculate that our new methods coupled with a more diverse training set (Figure ref{fig:trainset}) help our tools to outperform ONT base callers in such cases.

\begin{table*}
  \centering

  \begin{tabular}{|c|c|c|c|c|c|}
    \hline  Basecaller		& Mouse		& Rice 		& Zymoset		& \emph{Klebsiella}		& Speed [signals/s] \\
    \hline  Guppy 4.4 fast		& 85.887		& 89.146		& 91.569		& 91.466		& 512k \\
    \hline  Guppy 3.4 HAC		& 88.071		& 91.508		& \textbf{93.548}		& 93.588		& 55k \\
    \hline  Blitz 96		& 85.935		& 89.040		& 90.528		& 91.429		& 1.46M \\
    \hline  Osprey w/o dynpooling		& 88.444		& 92.843		& 92.943		& 93.509		& 1.76M \\
    \hline  Osprey with dynpooling		& \textbf{88.802}		& \textbf{93.465}		& 93.327		& \textbf{94.015}		& 1.45-1.88M  \\
    \hline
  \end{tabular}
  
  \caption{
    \emph{ \label{tab:accfast} Median read accuracy of fast base callers.
        The speed is measured on a desktop CPU i7-7700K using four cores.
        The speed with dynpooling varies due to changes in average pooling factor.}
  }
\end{table*}

  \begin{figure}
    \centering
  
    \includegraphics[width=0.8\textwidth]{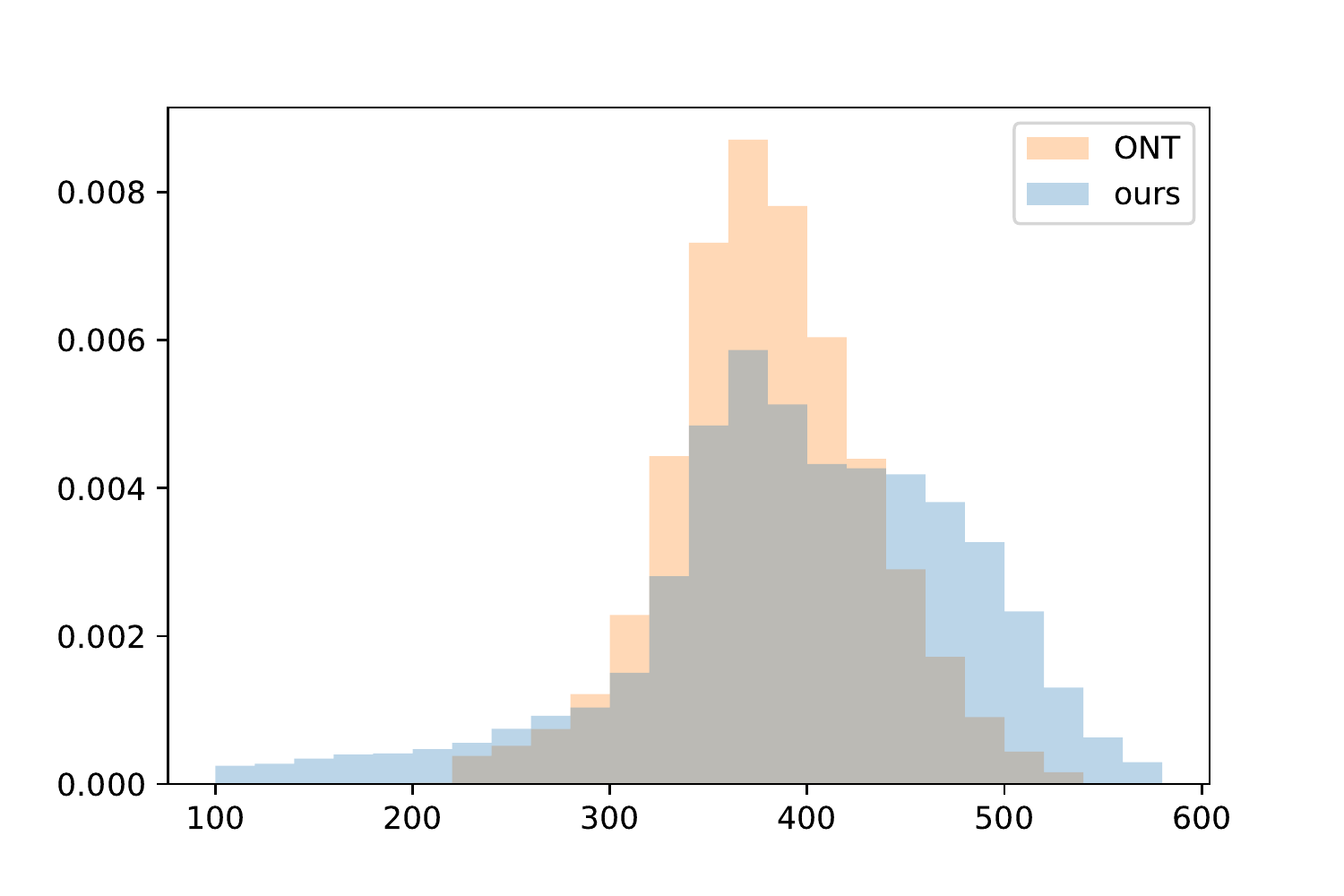}
    
    \caption{
      \emph{ \label{fig:trainset} Histogram of DNA speeds in training sets.}
    }
  \end{figure}

\subsection{Dynamic Pooling Effects}

\begin{figure}
    \centering
  
    \includegraphics[width=0.8\textwidth]{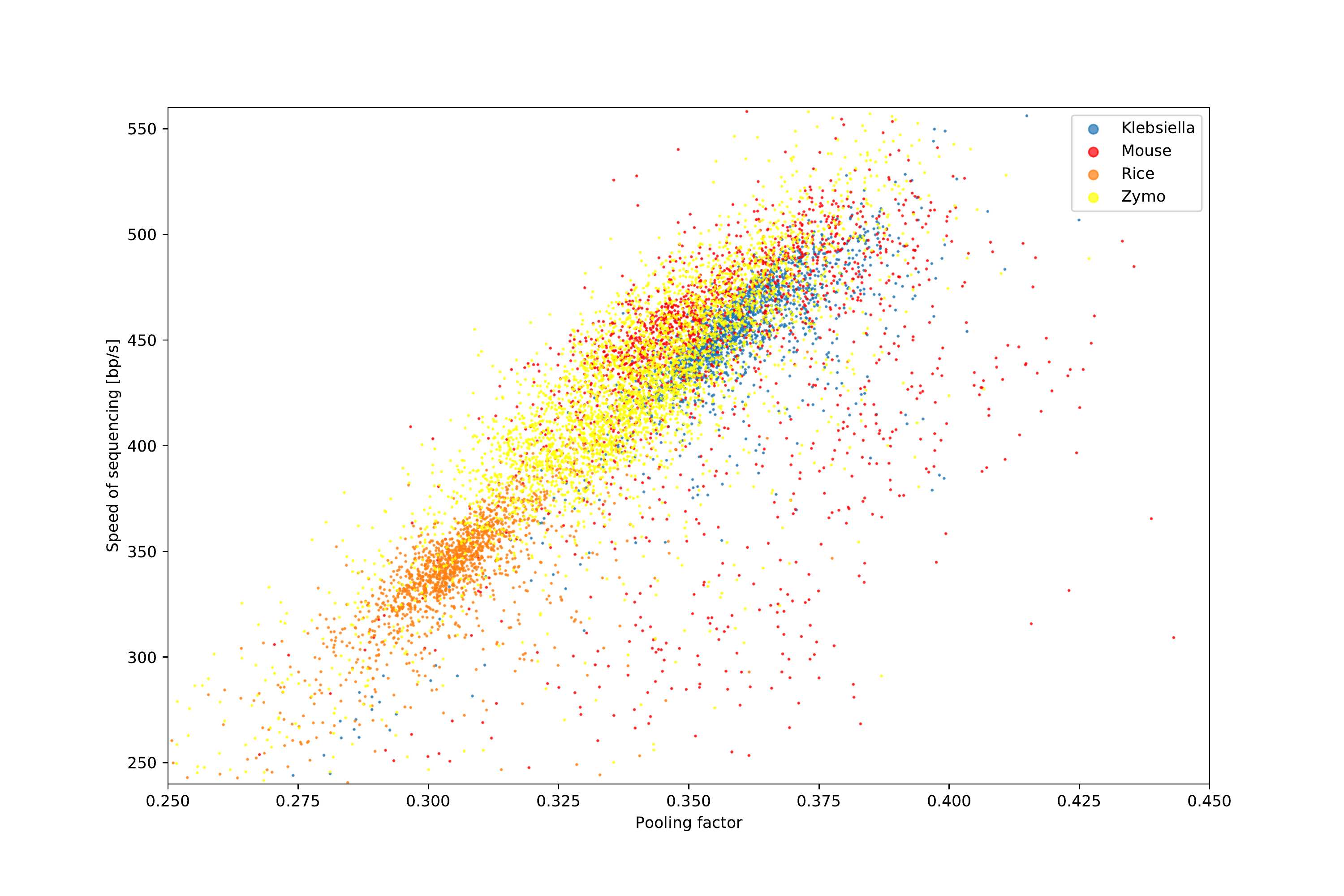}
    
    \caption{
      \emph{ \label{fig:dynpoolres} Correlation between the average pooling factor and the observed speed of sequencing calculated from the aligned reference length.
          Each dot corresponds to a single read and is colored by the dataset.}
    }
  \end{figure}

Figure \ref{fig:dynpoolres} shows that the dynamic pooling scheme indeed helps the network to adapt to varying sequencing speeds present in our data sets.
  In particular, there is a strong correlation between the average pooling factor and the observed DNA speed (computed as the ratio of the aligned sequence length to the sequencing time in seconds), with $r^2$ ranging from 0.41 for mouse to 0.82 for zymoset and \emph{Klebsiella} datasets.

\begin{figure}
    \centering
  
    \includegraphics[width=0.8\textwidth]{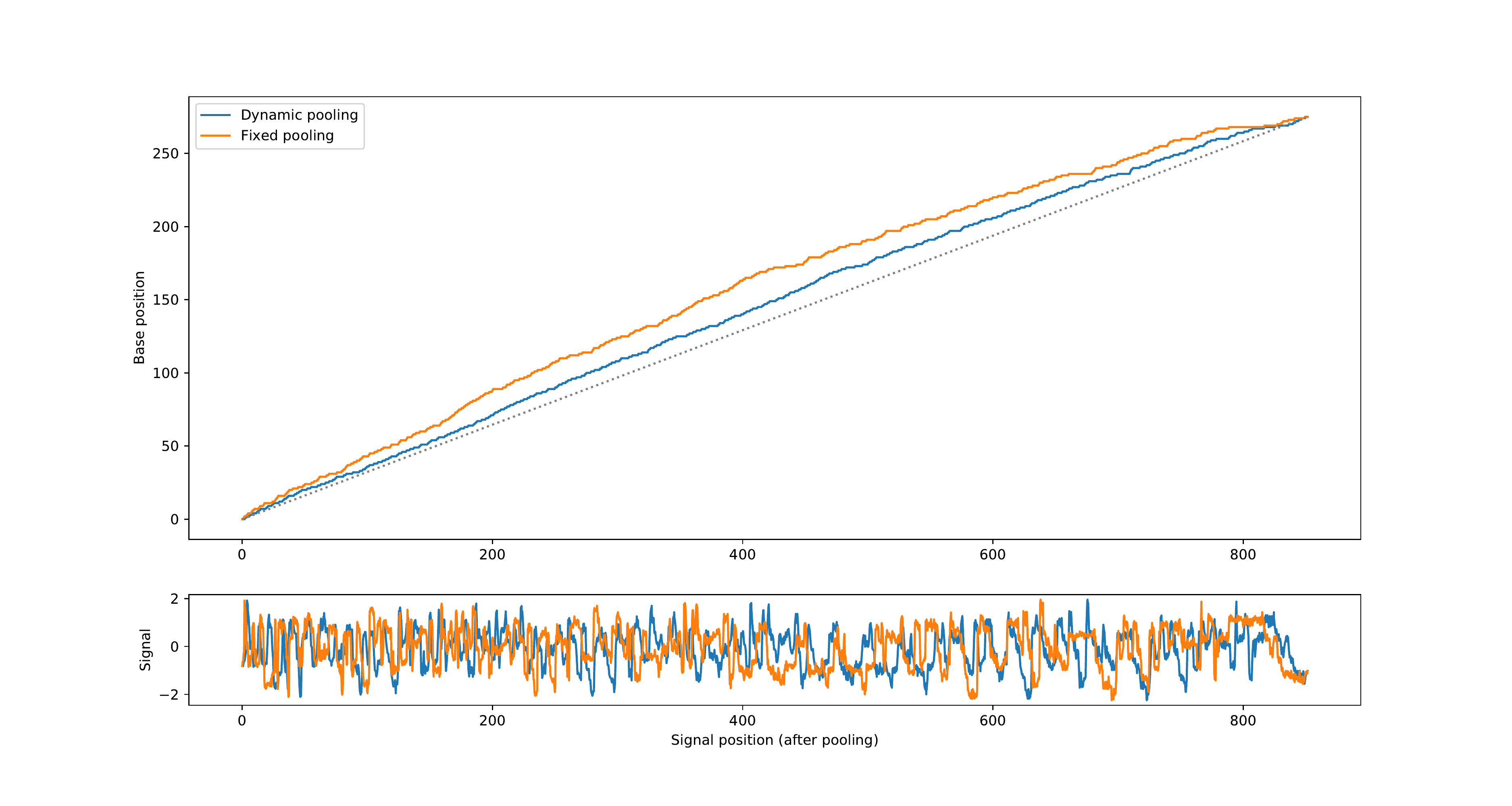}
    
    \caption{
      \emph{ \label{fig:dynpoolsignal1}  Detail of dynamic pooling vs fixed pooling on a single read.
          Top: Alignment between pooled signal and called bases.
          We see that dynamic pooling results in more uniform steps.
          Bottom: Actual signal mapped to positions after pooling.
          }
    }
  \end{figure}

\begin{figure}
    \centering
  
    \includegraphics[width=0.8\textwidth]{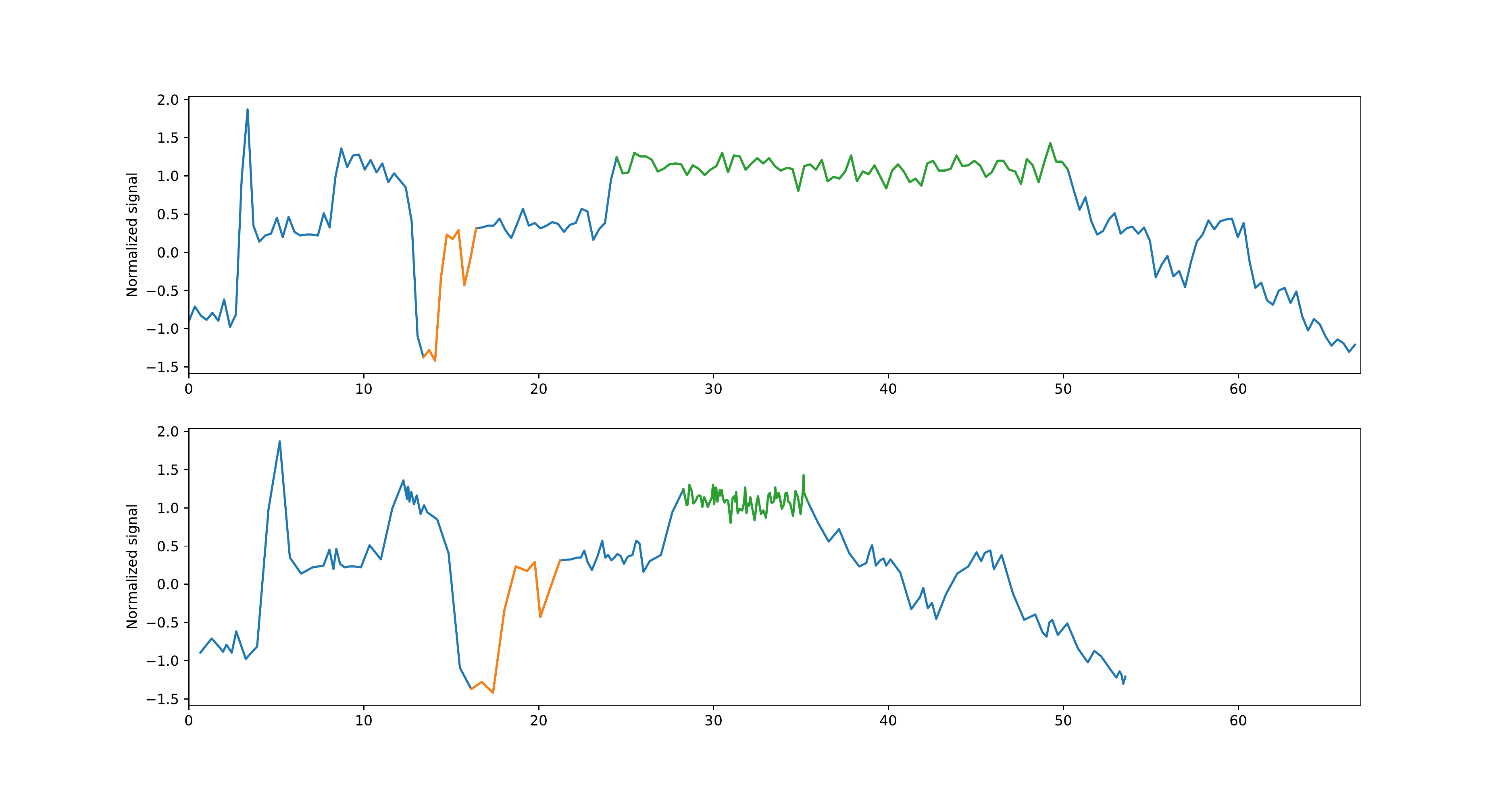}
    
    \caption{
      \emph{ \label{fig:dynpoolsignal2} A long event (green) is shortened by dynamic pooling, while several consecutive short events (orange)  are lengthened.
          Original signal is mapped to positions obtained by pooling.
          Top: Dynamic pooling.
          Bottom: Fixed pooling.}
    }
  \end{figure}

Figure \ref{fig:dynpoolsignal1} illustrates that dynamic pooling is able to compensate in part for the changes in sequencing speed within a read;
   figure \ref{fig:dynpoolsignal2} illustrates that long events can be shortened and short events lengthened.
  Overall we see that dynamic pooling results in a much more uniform number of signal steps per called base.

\section{Conclusions and Future Work}

We presented dynamic pooling, a novel component for neural networks, and applied it to the problem of nanopore base calling.
  Dynamic pooling allows the network to better accommodate speed variation in the input signal and thus improve accuracy without increasing computational costs.

To demonstrate the effectiveness of dynamic pooling, we have developed two base callers for nanopore data.
   Osprey is a fast base caller, which allows near real-time base calling without the need of a GPU, while keeping the accuracy high enough for most sequence analysis workflows, since its accuracy exceeds the standard Guppy high accuracy base caller from a year ago (which is not real-time without a powerful GPU).
  Real-time base calling enables sequencing run monitoring and selective sequencing \cite{payne2021readfish}.

The second base caller, Heron, pushes the boundary of base calling quality.
  Heron can deliver consistently high accuracy on a variety of datasets, in some cases improving on the most accurate base caller to date.

While we developed dynamic pooling with nanopore base calling in mind, it has a potential to improve various other sequence related tasks, where the input signal has variable density of information.
  Example domains include speech recognition and video stream processing.
   Another intriguing goal is to extend dynamic pooling to multiple dimensions to be used in the context of image processing.
  However, this is not a straightforward task, since calculating input point positions in multiple dimensions is much more complicated than the prefix sums.

\textbf{Acknowledgements.}
  This research was supported by grants from the Slovak Research Grant Agency VEGA 1/0458/18 (TV) and 1/0463/20 (BB), Slovak Research and Development Agency APVV-18-0239, Horizon 2020 research and innovation programme No 872539 (PANGAIA), integrated infrastructure programme ITMS2014:313011ATL7, and by an academic research grant from Google Cloud Platform (VB).
  We also gratefully acknowledge a hardware grant from NVIDIA.

\bibliographystyle{plain}
\bibliography{main}

\renewcommand\thetable{S\arabic{table}}
\renewcommand\thefigure{S\arabic{figure}}
\renewcommand\thesection{S\arabic{section}}

\title{
  Dynamic pooling improves nanopore base calling accuracy\\
  Supplementary Material
}

\maketitle

\setcounter{section}{0}
\section{Depth-to-Space Compression}

In our previous work \cite{peresini2020nanopore}, we devised an alternative block for neural networks, which trades a lower number of sequence points for a higher number of channels.
  This is achieved by compressing the signal with strided convolution at the start of a convolutional block and decompressing it via strided transposed convolution at the end of the block.
  In our work, we increase the number of channels by two and compress sequence length by factor 3. This substantially increases number of parameters (by factor of 2), while increasing amount of multiplication only modestly (by factor of 4/3) and surprisingly decreasing real computation time.

Our design in \cite{peresini2020nanopore} was specifically tailored for inference on a rather restrictive neural network accelerator,  but we found that its benefits are also visible on a GPU after small adjustments.
  Specifically, because depthwise convolutions are faster on GPUs, we keep the first depthwise convolution after compression that we originally omitted from the block.

\begin{figure}
    \centering
  
    \includegraphics[width=0.8\textwidth]{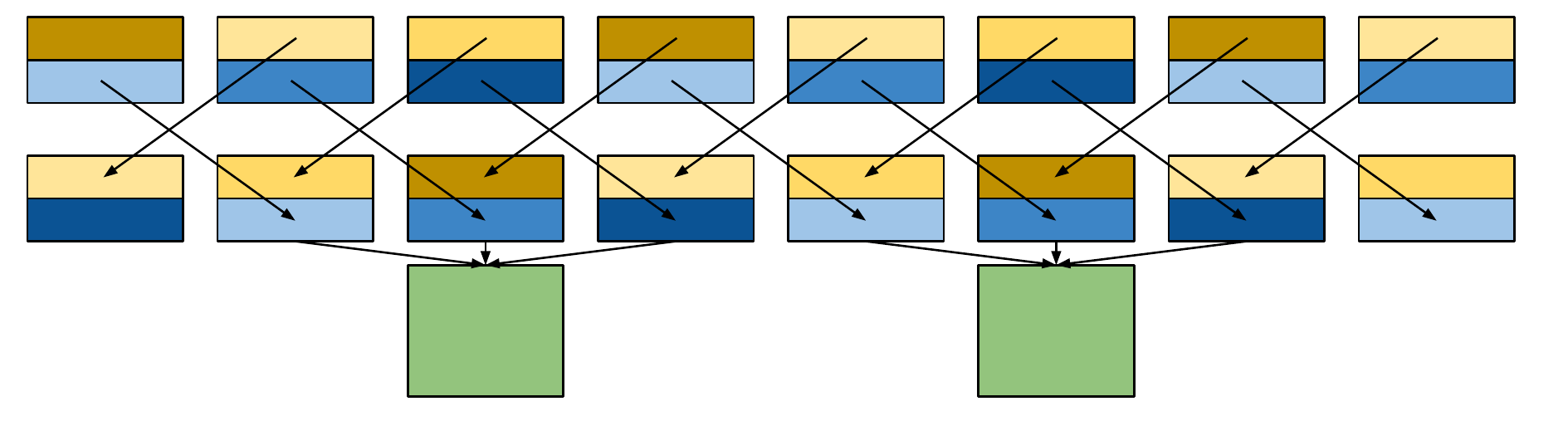}
    
    \caption{
      \emph{ \label{fig:shift} Visualization of the cross-shift operation followed by compression.}
    }
  \end{figure}

In the Heron high accuracy network, we also introduce a novel ``channel-cross-shift'' operation before the compression, which is designed to increase communication between compressed parts of the sequence.
  We split channels in each sequence point into two halves A and B, and shift A by one point to the right and B by one point to the left, with empty slots padded with zeroes (Figure \ref{fig:shift}).
  In practice, while the cross-shift does not seem to significantly improve the overall accuracy after convergence, it improves the training curve \ref{fig:training}.

The number of operations in compression and decompression is still high for Osprey fast CPU implementation (both operations are equivalent to two pointwise convolutions per sequence point).
  Surprisingly we can simplify compression to average pooling followed by pointwise convolution, which effectively decreases the number of operations by two thirds during compression and does not hurt the accuracy much (when keeping the same number of floating point operations, this design is actually slightly better).
  Simplifying decompression is not so easy.
  Here we opt for a rather arbitrary design where we split channels into overlapping groups and each group is feeded into one pointwise convolution, which produces one third of the outputs.

\section{Gated linear units }

Many modern convolutional neural networks, including Bonito v0.2, use Swish activation function $f(x)=x\sigma (x)$ \cite{ramachandran2017searching}.
  An interesting extension, which comes at the cost of doubling the amount of computation, is to use a gated linear unit (GLU) \cite{dauphin2017language}, which first splits channels into two halves ($x_1,x_2$) and then applies transformation $g(x_1,x_2)=x_1\sigma (x_2)$.
  Note that Swish is a special case of GLU with$x_1=x_2$ .

We use GLUs in Heron high accuracy network, doubling the output size of all pointwise transformations to compensate for GLUs cutting the number of channels by half.
  A FLOPs equivalent of this new GLU architecture would be the original Swish architecture with the number of filters in the whole network increased by factor 1.41.
   Interestingly, when we tried to train such a large Swish network, it did not bring any increase in accuracy (see \ref{tab:accfull}).
  As such we think that either the GLU network benefits from the more complex channel interactions, or that the large Swish network fails to train properly.
  Also note that introduction of GLUs does not lead on a GPU to the expected doubling of the running time, perhaps because more time is spent in depthwise convolutions, and increasing the size of pointwise convolutions does not hurt the speed that much.

Finally, we observed that using standard initialization from Pytorch did not work for our network.
  Instead, following the methodology from \cite{merity2019single} chapter 6.4.1, we devised a custom initialization scheme.
  We split all weight matrices before GLU nonlinearity by half (in the output dimension) and we initialize only first half of the matrix randomly and set second half to the same values as the first half.
  Due to this initialization the initial state of the network mimics network using Swish nonlinearity.

\section{Recurrent Neural Aligner}

  \begin{figure}
    \centering
  
    \includegraphics[width=0.7\textwidth]{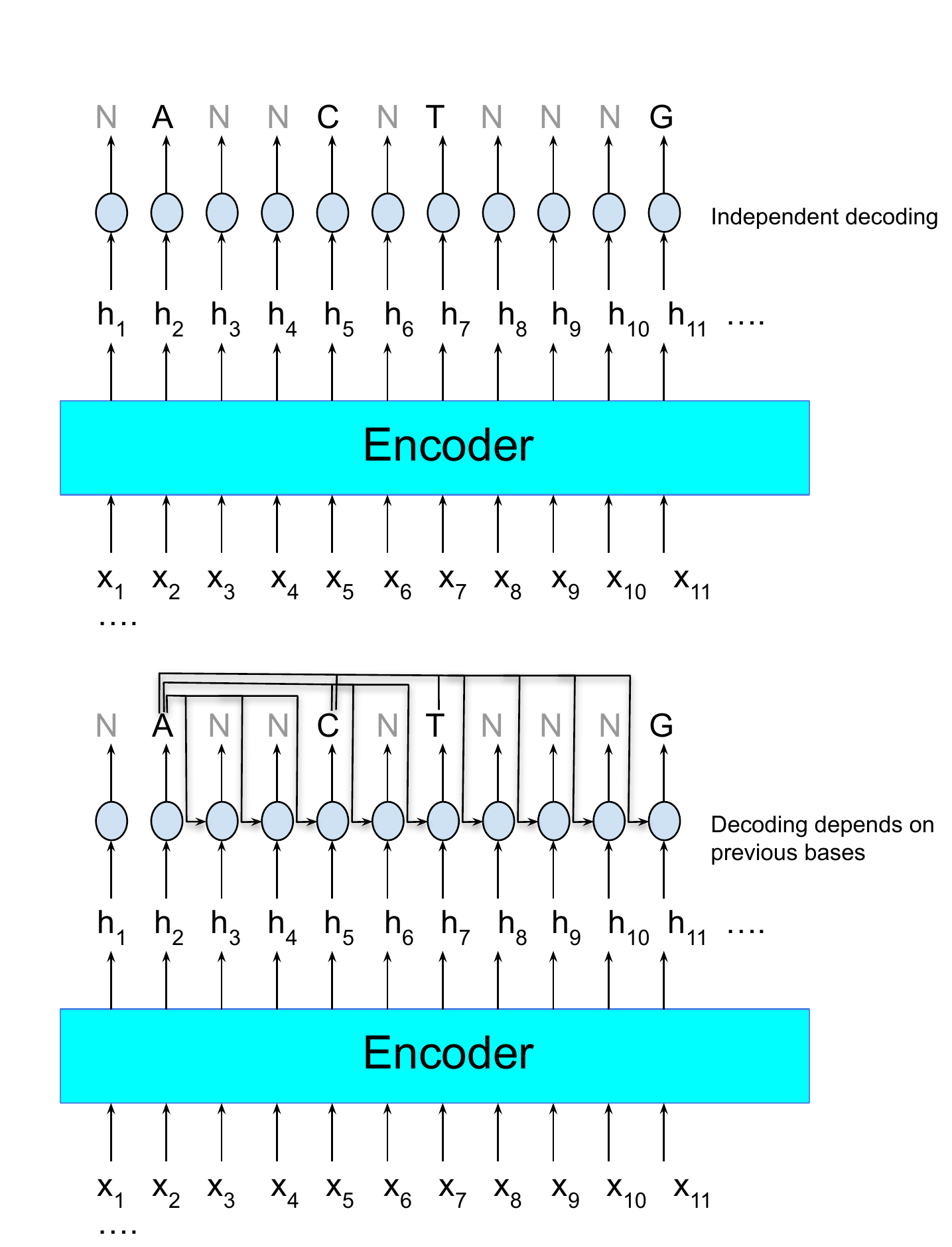}
    
    \caption{
      \emph{ \label{fig:trans_arch} Illustration of probabilistic models underlying the CTC and recurrent neural aligner frameworks.
          To sample a sequence $y_1,\dots, y_n$ of intermediate predictions from alphabet $\Sigma_\epsilon$ in CTC, each $y_i$ would be generated independently (top), while in the recurrent neural aligner, the probability of $y_i$ would depend on the previous non-blank symbols (bottom).
          Note that the actual prediction of the network is a sequence of non-blank symbols with the highest probability, so the sampling is not used in the training and inference algorithms.}
    }
  \end{figure}

The last stage of base calling, called decoder, is responsible for converting the sequence of high-level features predicted by the CNN into the final DNA sequence.
  Because typically multiple signal points correspond to the same DNA base, many base callers (e.g.
  Deepnano-blitz, Chiron, Bonito v0.2) use the connectionist temporal classification (CTC) decoder \cite{graves2006connectionist}.
  The network outputs a probability distribution over the five-value alphabet $\Sigma_\varepsilon = \{A,C,G,T,\varepsilon\}$ at each signal point.
  This defines a probability $p_Y$ of observing sequence $Y = y_1,\dots, y_n$ over the alphabet $\Sigma_\varepsilon$ as the product of point probabilities of individual characters $y_i$.

Let $R(Y)$ be a function that reduces such a sequence $Y$ to the final output by omitting blank symbols $\varepsilon$ (this function can be alternatively defined to perform additional operations, such as replacing runs of identical symbols to a single symbol).
  Note that several different sequences $Y$ may yield the same reduced sequence $Z$.
  The CTC decoder aims to find the output $Z$ with the highest overall probability $\arg\max_Z \sum_{Y:R(Y)=Z} p_Y$.
  Such a layer is differentiable and can be trained by standard algorithms based on gradient descent.
  However, inference of the best sequence $Z$ in a trained network is hard, and consequently beam search heuristic is typically used.

The underlying probabilistic model of CTC assumes that point predictions $y_i$ are independent of each other.
  This simplifying assumption enables further applications such as pair-decoding \cite{silvestre2020pair}.
  However, a more complex model with dependencies between positions may simplify the necessary computations within the main CNN.
  For this purpose we use a recurrent neural aligner framework \cite{sak2017recurrent}.

In a recurrent neural aligner, the probability distribution at sequence point $i$ depends not only on the input but also on the previous predictions (Figure \ref{fig:trans_arch}).
  More precisely, the probability $q_Y$ of $Y = y_1,\dots, y_n$ over the alphabet $\Sigma_\varepsilon$ is the product of conditional distributions
$$q_Y = \prod_{i=1}^n \Pr(y_i\,|\,h_i, R(y_1, \dots, y_{i-1})),$$
where $h_i$is the $i$-th output of the CNN.
  The conditional probability distribution $Pr(y_i\,|\,h_i, R(y_1, \dots, y_{i-1})$ for all five values $y_i\in \Sigma_\varepsilon$ will be computed by some mapping $Q(h_i, z_1, \dots z_j)$, which gets as an input vector $h_i$ and string $z_1, \dots z_j = R(y_1, \dots, y_{i-1})$.
  This mapping can be an arbitrary differentiable function producing valid probability distributions.

Similarly, as in CTC, the probability of output sequence $Z$ is simply $\Pr(Z\,|\,h_1, \dots, h_n) = \sum_{Y:R(Y)=Z} q_Y$.
  Then the model can be trained by back propagation via log likelihood loss.
  To compute the likelihood of a known ground-truth output Z=$z_1,z{\ }_{2,}...,z_m$, we first compute conditional probability distributions $Q(h_i, z_1, \dots, z_j)$ for every pair of i, j ($i\geq j)$.
  The final likelihood of $Z$ can be then computed via dynamic programming using the forward algorithm.
  During inference, we use the beam search where we keep the top $k$ candidates for each prefix of the predicted sequence.
  Compared to CTC, where beam sizes of 10 are usually sufficient, we need to use a much higher beam size with recurrent neural aligners (we use $k=50$).

Finally, we describe the design of conditional probability distribution $Q(h_i,z{\ }_1,...,z{\ }_j)$.
  We split its calculation into two parts.
  We first embed the previous non-blank predictions $z_1, \dots z_j$ into a fixed-size vector via prediction encoder $G$.
  We then mix $h_i$and the output of $G$ via function $H$ as  $Q(h_i, z_1,\dots, z_j)= H(h_i,G(z_1, \dots, z_j))$.

We tested several variants for the design of functions $G$ and $H$.
   One simple option is to consider only the last $k$ predictions $z_{j-k+1}, \dots, z_j$.
  In our work, we use $k=6$.
  Using just the last several predictions is motivated by the fact that the signal current depends mostly on the small context around the current base in the pore.
  Then $G$ can be a table containing for each of $4^k$ input strings a small matrix $W$ of trained weights and a bias vector $b$.
  Function H will facilitate vector-matrix multiplication between $h_i$and the result of G as follows $Wh\ +b$.
  The resulting vector of length five is transformed by the softmax function to produce a proper probability distribution.

Another option is to use RNN for $G$, which encodes the sequence of bases into a fixed size vector at each position and then e.g.
  use $H(h,g)=W_2(h*ReLU(W_1g\ +\ b_1))\ +b_2$.
  We find that this design does not lead to a significant increase in base calling accuracy and causes decoding to become much slower.
  Also if the dimension of $h$ is small, we find that training sometimes depends on luck during initialization.

\section{Training methodology}

\begin{figure}
    \centering
  
    \includegraphics[width=0.8\textwidth]{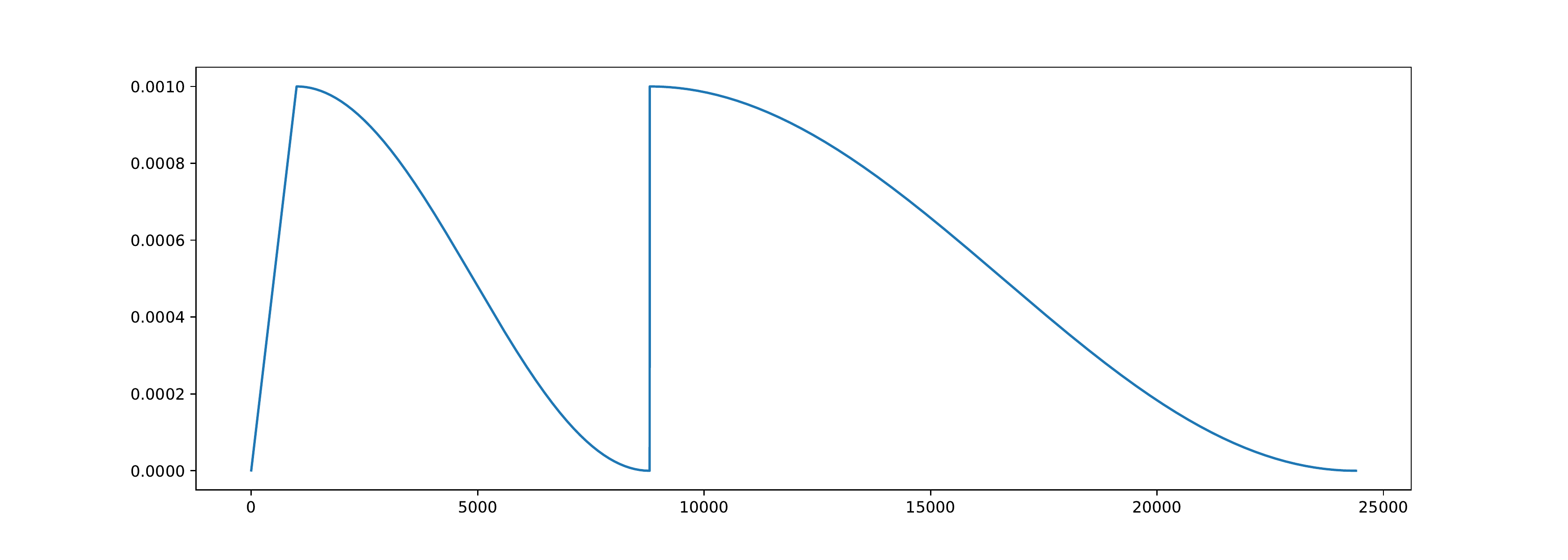}
    
    \caption{
      \emph{ \label{fig:lr} Learning rate schedule for the first two cycles.}
    }
  \end{figure}

We train all of our high accuracy networks with AdamW (\cite{loshchilov2018decoupled}) algorithm.
  We use a cyclic cosine learning rate scheduler with warm restarts (Figure \ref{fig:lr}) \cite{loshchilov2017sgdr}.
  We anneal the learning rate from maximum of 1e-3 toward zero, and we exponentially increase the length of cycles by a factor of 2 for each successive cycle.
  This results in multiple possible stopping points for training (where each of them has converged with a small learning rate).
  We used a fixed budget of 7 cycles (which means 127 multiples of the first cycle in total).
  The first cycle takes 8000 batches.
  We start our learning by a quick warmup period for 1000 batches.

\begin{figure}
    \centering
  
    \includegraphics[width=0.8\textwidth]{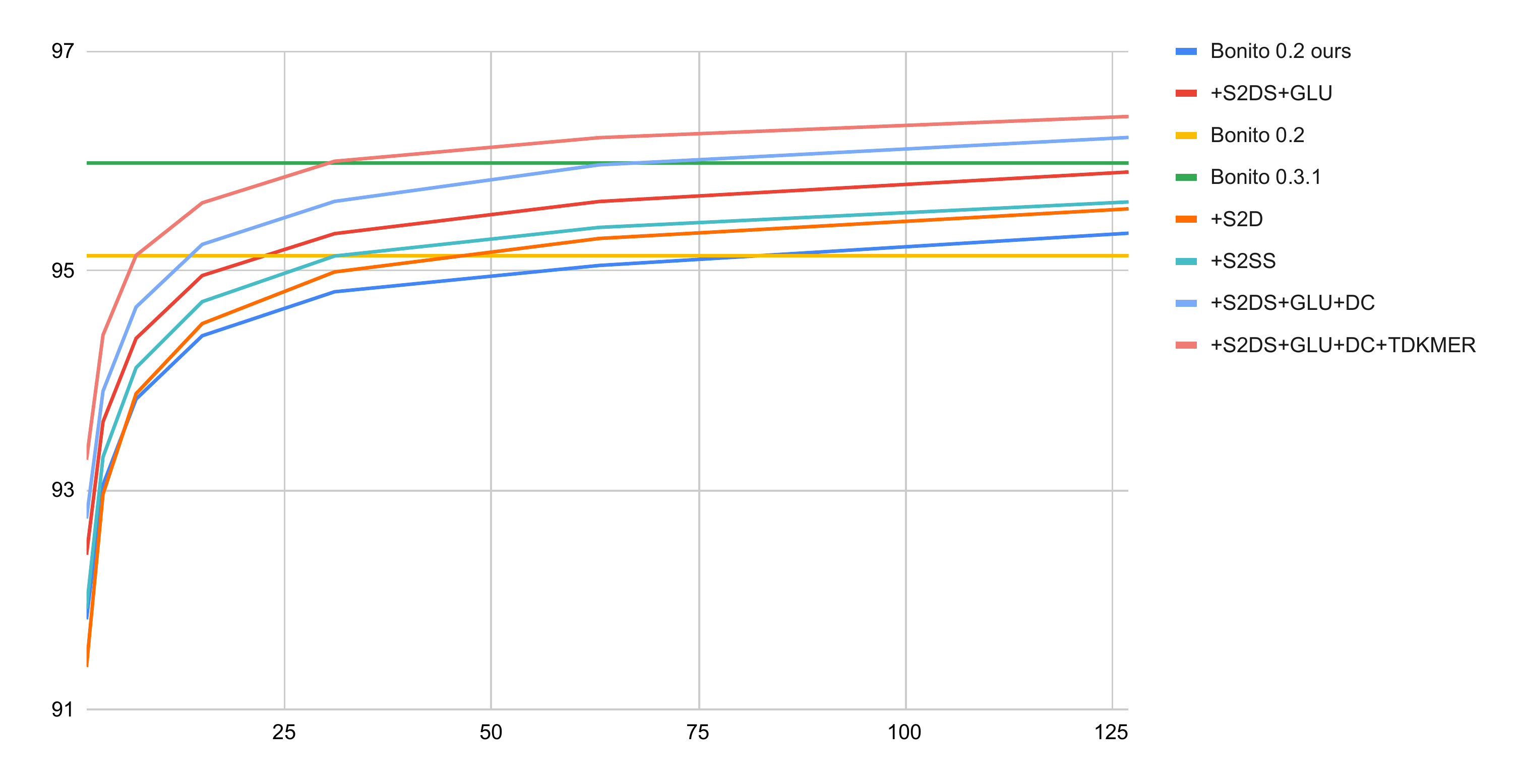}
    
    \caption{
      \emph{ \label{fig:training} Convergence of various models over time.}
    }
  \end{figure}

Based on the convergence plot \ref{fig:training}, we can see that our models almost converged, but there is still a small room for more optimization.
  However, the last cycle of training took around 4-6 days (depending on the network) over 2-4 GPUs.
  We decided that taking double of that computational budget is not worth a small increase in performance.

We use the same setup for fast models, but we increase the batch size by a factor of 5, so we can more efficiently utilize the GPUs.
  We also use 5 times fewer steps for each cycle and use 5 times higher learning rate.

\section{Ablation Results}

We also measured the effect of individual improvements in Heron base caller.
  We started by retraining Bonito 0.2 architecture on our training dataset with our optimizer settings and iteratively added more improvements to it.
   Table \ref{tab:accfull} shows that each improvement individually brings only a modest gain but together the gain is quite substantial.
  Surprisingly, increasing the number of channels (to match FLOPs equivalent of enlarging the network due to GLU activation) leads to degradation of performance.
  This was not due to overfitting, but actually due to worse training performance.
  We hypothesize that this was caused by bad initialization of the network.

\begin{table*}[h]
  \centering

  \begin{tabular}{|c|c|c|c|c|c|}
    \hline  Base caller		& Mouse		& Rice 		& Zymoset		& Klebsiella		& Speed [signals/s] \\
    \hline  Bonito 0.3.1		& 90.770		& 93.477		& \textbf{96.435}		& 95.981		& 0.37M \\
    \hline  Bonito 0.3		& \textbf{91.233}		&         94.129		& 95.733		& 95.914		& 0.37M \\
    \hline  Bonito 0.2		& 89.489		& 93.040		& 94.999		& 95.136		& 0.44M \\
    \hline  Bonito 0.2 ours		& 90.142		& 94.885		& 94.919		& 95.343		& 0.50M \\
    \hline  +S2D		& 90.190		& 95.328		& 95.151		& 95.565		& 0.65M \\
    \hline  +S2DS		& 90.285		& 95.376		& 95.210		& 95.628		& 0.65M \\
    \hline  \emph{+S2DS with 1.5x channels}		& \emph{90.012}		& \emph{95.017}		& \emph{94.883}		& \emph{95.334}		& \emph{0.32M} \\
    \hline  +S2DS+GLU		& 90.611		& 95.551		& 95.412		& 95.901		& 0.45M \\
    \hline  +S2DS+GLU+DP		& 90.886		& 95.928		& 95.726		& 96.216		& 0.41M \\
    \hline  +S2DS+GLU+TD		& 90.633		& 95.822		& 95.596		& 96.037		& 0.45M \\
    \hline  +S2DS+GLU+DP+TD		& 90.911		& 96.092		& 95.907		& \textbf{96.407}		& 0.41M \\
    \hline
  \end{tabular}
  
  \caption{
    \emph{ \label{tab:accfull} Median accuracy of various variants of high accuracy basecallers (we measure effect on improvements over bonito 0.2 architecture).
        Legend: S2D - space to depth compression.
        S2DS - space to depth compression with cross-shift.
        GLU - gated linear units.
        DP - dynamic pooling, TD - transducer (recurrent neural aligner)}
  }
\end{table*}

\end{document}